\documentclass[11pt,a4paper]{article}
\usepackage{times,latexsym}
\usepackage{url}
\usepackage[T1]{fontenc}
\usepackage[]{style/emnlp-ijcnlp-2019}
\usepackage{style/mystyle}
\usepackage{style/symbol}
\usepackage{tikz}
\usepackage{float}
\usepackage{arydshln}
\usepackage{centernot}
\usepackage{bbm}
\usepackage{float}

\aclfinalcopy 

\newcommand{\theTitle}{A Logic-Driven Framework for Consistency of Neural Models}

\title{\theTitle}

\author{Tao Li, Vivek Gupta, Maitrey Mehta, Vivek Srikumar\\
School of Computing, University of Utah\\
\tt{\{tli,vgupta,maitrey,svivek\}@cs.utah.edu}
}

\date{}

\begin{document}
\maketitle

\begin{abstract}
  While neural models show remarkable accuracy on individual predictions, their
  internal beliefs can be inconsistent \emph{across} examples.  In this paper,
  we formalize such inconsistency as a generalization of prediction error. We
  propose a learning framework for constraining models using logic rules to
  regularize them away from inconsistency. Our framework can leverage both
  labeled and unlabeled examples and is directly compatible with off-the-shelf
  learning schemes without model redesign.  We instantiate our framework on
  natural language inference, where experiments show that enforcing invariants
  stated in logic can help make the predictions of neural models both accurate
  and consistent.

\end{abstract}


\section{Introduction}
\label{sec:intro}

Recent NLP advances have been powered by improved representations~\cite[\eg,
ELMo, BERT---][]{peters2018deep,devlin2018bert}, novel neural
architectures~\cite[\eg,][]{cheng2016long,seo2016bidirectional,parikh2016decomposable,vaswani2017attention},
and large labeled
corpora~\cite[\eg,][]{bowman2015large,rajpurkar2016squad,N18-1101}.
Consequently, we have seen progressively improving performances on benchmarks
such as GLUE~\cite{wang2018glue}. But, are models really becoming better?
We take the position that, while tracking performance on a leaderboard is
necessary to characterize model quality, it is not sufficient. Reasoning about
language requires that a system has the ability not only to draw correct inferences
about textual inputs, but also to be consistent its beliefs across various inputs.

To illustrate this notion of consistency, let us consider the task of natural
language inference (NLI) which seeks to identify whether a premise entails,
contradicts or is unrelated to a hypothesis~\cite{dagan2013recognizing}. Suppose
we have three sentences $P$, $H$ and $Z$, where $P$ entails $H$ and $H$
contradicts $Z$. Using these two facts, we can infer that $P$ contradicts
$Z$. In other words, these three decisions are not independent of each
other. Any model for textual inference should not violate this invariant defined
over any three sentences, \emph{even if they are not labeled}.

Neither are today's models trained to be consistent in this fashion, nor is
consistency evaluated. The decomposable attention model
of~\citet{parikh2016decomposable} updated with ELMo violates the above
constraint for the following sentences:\footnote{We used the model available
  through the Allen NLP online demo:
  \url{http://demo.allennlp.org/textual-entailment}.}
\begin{itemize}[noitemsep]
\item[$P$:] John is on a train to Berlin.
\item[$H$:] John is traveling to Berlin.
\item[$Z$:] John is having lunch in Berlin.
\end{itemize}
Highly accurate models can be inconsistent in their beliefs over
groups of examples. For example, using a BERT-based NLI model that
achieves about 90\% F-score on the SNLI test
set~\cite{bowman2015large}, we found that in about 46\% of {\em
  unlabeled} sentence triples where $P$ entails $H$ and $H$
contradicts $Z$, the first sentence does not contradict the
third. Observations of a similar spirit were also made by
\citet{minervini2018adversarially}, \citet{glockner2018breaking} and \citet{nie2018analyzing}.

To characterize and eliminate such errors, first, we define a method to measure
the inconsistency of models with respect to invariants stated as first-order
logic formulas over model predictions. We show that our definition of
inconsistency strictly generalizes the standard definition of model error.



Second, we develop a systematic framework for mitigating inconsistency in models
by compiling the invariants into a differentiable loss functions using
t-norms~\cite[][]{klement2013triangular,gupta1991theory} to soften logic. This allows us to take
advantage of unlabeled examples and enforce consistency of model predictions
over them.  We show that the commonly used cross-entropy loss emerges as a
specific instance of our framework.  Our framework can be easily instantiated
with modern neural network architectures.



To show the effectiveness of our approach, we instantiate it on the NLI task.
We show that even state-of-the-art models can be highly inconsistent in their
predictions, but our approach significantly reduces inconsistency.

In summary, our contributions are:
\begin{enumerate}[nosep]
\item We define a mechanism to measure model inconsistency with respect to
  declaratively specified invariants.
\item We present a framework that compiles knowledge stated in first-order logic
  to loss functions that mitigate inconsistency.
  \item We show that our learning framework can reduce prediction
    inconsistencies even with small amount of annotated examples
    without sacrificing predictive accuracy.\footnote{Our code to replay our
      experiments is archived at \url{https://github.com/utahnlp/consistency}.}
\end{enumerate}

\section{A Framework for (In)consistency}
\label{sec:framework}
In this section, we will present a systematic approach for measuring
and mitigating inconsistent predictions.
%
A prediction is {\em incorrect} if it disagrees with what is known to be true. 
Similarly, predictions are {\em inconsistent} if they do not follow a known rule.
Therefore, a model's errors can
be defined by their concordance with declarative knowledge.
We will formalize this intuition by first developing a uniform representation
for both labeled examples and consistency constraints
(\S\ref{sec:representing-knowledge}).  Then, we will present a general
definition of errors in the context of such a representation
(\S\ref{sec:error-as-inconsistency}).  Finally, we will show a logic-driven
approach for designing training losses (\S~\ref{sec:deriving-loss-functions}).

As a running example, we will use the NLI task whose goal is to predict one of
three labels: \emph{Entailment} ($E$), \emph{Contradiction} ($C$), or
\emph{Neutral} ($N$).



\subsection{Representing Knowledge}
\label{sec:representing-knowledge}
Suppose $x$ is a {\em collection} of examples (perhaps labeled). We write
constraints about them as a conjunction of statements in logic:
\begin{align}
  \quad \Land_{(L,R)} L(x) \rightarrow R(x) \label{eq:top}
\end{align}
Here, $L$ and $R$ are Boolean formulas, \ie antecedents and consequents,
constructed from model predictions on examples in $x$.

One example of such an invariant is the constraint from \S\ref{sec:intro}, which
can be written as $E(P, H) \land C(H, Z) \rightarrow C(P, Z)$, where, \eg,
predicate $E(P,H)$ denotes that model predicted label $E$.  We can also
represent labeled examples as constraints: \emph{``If an example is annotated
  with label $Y^\star$, then model should predict so.''}  In logic, we write
$\top \rightarrow Y^\star(x)$.\footnote{The symbol $\top$ denotes the Boolean
  {\tt true}.}  Seen this way, the expression~\eqref{eq:top} could represent
labeled data, unlabeled groups of examples with constraints between them, or a
combination.

\subsection{Generalizing Errors as Inconsistencies}
\label{sec:error-as-inconsistency}
Using the representation defined above, we can define how to evaluate predictors.
We seek two properties of an evaluation metric: It should
\begin{inparaenum}[1)]
\item quantify the inconsistency of predictions, and
\item also generalize classification error.
\end{inparaenum}
To this end, we define two types of errors: {\em global} and {\em conditional} violation.
Both are defined for a dataset $D$ consisting of example collections $x$ as
described above.

%

\paragraph{Global Violation ($\boldsymbol \rho$)}
The global violation is the fraction of examples in a dataset $D$
where any constraint is violated. We have:
\begin{align}
    \rho &= \frac{\Sum_{x\in D} \pb{\Lor_{(L,R)} \neg \p{L(x) \rightarrow R(x)}}}{|D|} \label{eq:global}
\end{align}
Here, $\pb{\cdot}$ is the indicator function.

\paragraph{Conditional Violation ($\boldsymbol \tau$)} For a
conditional statement, if the antecedent is not satisfied, the
statement becomes trivially true. Thus, with complex antecedents,
the number of examples where the constraint is true can be trivially large.
To only consider those examples where the antecedent holds,
we define the conditional violation as:
\begin{align}
    \tau &= \frac{\Sum_{x\in D} \pb{\Lor_{(L,R)} \neg \p{L(x) \rightarrow R(x)}}}{\Sum_{x\in D} \pb{\Lor_{(L,R)} L(x)}}    \label{eq:conditional}
\end{align}

\paragraph{Discussion}

The two metrics are complementary to each other.
On one hand, to lower the global metric $\rho$, a model could avoid satisfying the antecedents.
In this case, the conditional metric $\tau$ is more informative.
On the other hand, the global metric reflects the impact of domain
knowledge in a given dataset, while the conditional one does not.
Ideally, both should be low.

Both violations strictly generalize classification error.  If all the knowledge
we have takes the form of labeled examples, as exemplified at the end of
\S\ref{sec:representing-knowledge}, both violation metrics are identical to
model error. The appendix formally shows this.

%
\begin{table*}[ht]
  \centering
  \begin{tabular}{lcccc}
    \toprule
    Name & Boolean Logic          & Product & G\"odel & \L{}ukasiewicz \\
    \midrule    
    Negation                      & $\neg A$     & $ 1-a $ & $1-a$          & $1-a$             \\
    T-norm & $A \wedge B$ & $ab$    & $\min\p{a, b}$ & $\max\p{0,a+b-1}$ \\
    T-conorm & $A \vee B$   & $a+b-ab$ & $\max(a, b)$ & $\min\p{1,a+b}$   \\
    Residuum  & $A \rightarrow B$ & $\min\p{1, \frac{b}{a}}$ & $\begin{cases}
      1, \text{ if b $\geq$ a},\\
      b, \text{ else} \\
    \end{cases}$ & $\min\p{1, 1-a+b}$\\
    \bottomrule
  \end{tabular}
  \caption{Mapping discrete statements to differentiable functions using t-norms.
  Literals are upper-cased (e.g. $A$) while real-valued probabilities are lower-cased (e.g. $a$).
  Here, differentiable forms are from a mixture of $R$-fuzzy logic and $S$-fuzzy logic.
  In this paper, we focus on the product t-norm.}
  \label{tab:tnorm}
\end{table*}


\subsection{Learning by Minimizing Inconsistencies}
\label{sec:deriving-loss-functions}

With the notion of errors, we can now focus on how to train models to minimize them.
A key technical challenge involves the unification of discrete declarative
constraints with the standard loss-driven learning paradigm.

%

To address this, we will use relaxations of logic in the form of
t-norms to deterministically compile rules into differentiable loss
functions.\footnote{A full description of t-norms is beyond the
  scope of this paper; we refer the interested reader to
  \citet{klement2013triangular}.}
We treat predicted label probabilities as soft surrogates for Boolean decisions.
In the rest of the paper, we will use lower case for model probabilities---\eg, $e(P,H)$,
and upper case---\eg, $E(P, H)$---for Boolean predicates.

Different t-norms map the standard Boolean operations into different continuous
functions. Table~\ref{tab:tnorm} summarizes this mapping for three t-norms: product, G\"odel, and
\L{}ukasiewicz.
Complex Boolean expressions can be constructed from these four operations.
Thus, with t-norms to relax logic, we can systematically convert rules as in
\eqref{eq:top} into differentiable functions, which in turn serve as learning objectives to minimize constraint violations.
We can use any off-the-shelf optimizer~\cite[\eg, ADAM][]{kingma2014adam}.
We will see concrete examples in the NLI case study in \S\ref{sec:case}.

Picking a t-norm is both a design choice and an algorithmic one. Different
t-norms have different numerical characteristics and their comparison is a
question for future research.\footnote{For example, the G\"odel t-norm, used by
  \citet{minervini2018adversarially}, has a discountinuous but
  semi-differentiable residuum. The \L{}ukasiewicz t-norm can lead to zero
  gradients for large disjunctions, rendering learning difficult.}  Here, we
will focus on the product t-norm to allow comparisons to previous work: as we
will see in the next section, the product t-norm strictly generalizes the widely
used cross entropy loss.

\section{Case Study: NLI}
\label{sec:case}
We study our framework using the NLI task as a case study. First, in
\S\ref{sec:nli-learning-objective}, we will show how to represent a training set
as in \eqref{eq:top}. We will also introduce two classes of domain constraints
that apply to groups of premise-hypothesis pairs. Next, we will show how to
compile these declaratively stated learning objectives to loss functions
(\S\ref{sec:inconsistency-losses}). Finally, we will end this case study with a
discussion about practical issues (\S\ref{sec:nli-training}).

\subsection{Learning Objectives in Logic}
\label{sec:nli-learning-objective}
Our goal is to build models that minimize inconsistency with domain
knowledge stated in logic. Let us look at three such consistency
requirements.

\paragraph{Annotation Consistency}
For labeled examples, we expect that a model should predict what an annotator
specifies. That is, we require
\begin{align}
   \forall {(P,H),Y^\star \in D},\quad \top \rightarrow Y^\star{\p{P,H}}   \label{eq:labeled}
\end{align}
where $Y^\star$ represents the ground truth label for the example
$(P, H)$.
As mentioned at the end of \S\ref{sec:error-as-inconsistency}, for the
annotation consistency, both global and conditional violation rates
are the same, and minimizing them is maximizing accuracy.  In our
experiments, we will report accuracy instead of violation rate for
annotation consistency (to align with the literature).

\paragraph{Symmetry Consistency}
Given any premise-hypothesis pair, the grounds for a model to predict \emph{Contradiction} is that the events in the premise and
the hypothesis cannot coexist simultaneously.
That is, a $(P, H)$ pair is a contradiction if, and only if, the $(H,P)$ pair is also a contradiction:
\begin{align}
   \forall{(P,H) \in D},\quad C{\p{P,H}} \leftrightarrow C{\p{H,P}}   \label{eq:pairwise}
\end{align}

\paragraph{Transitivity Consistency} 
This constraint is applicable to any three related sentences $P$,
$H$ and $Z$. If we group the sentences into three pairs, namely
$(P,H)$, $(H, Z)$ and $(P,Z)$, the label definitions mandate that not all of the
$3^3 = 27$ assignments to these three pairs are allowed.
The example in \S\ref{sec:intro} is an allowed label assignment.
We can enumerate all such valid labels as the conjunction:
\begin{align}
    \begin{split}
    \forall{(P,H,Z) \in D},\quad\quad\quad\quad \\
    \left(E\p{P,H} \wedge E\p{H,Z}\right. &  \rightarrow	\left.E\p{P,Z}\right) \\
    \land \left(E\p{P,H} \wedge C\p{H,Z}\right. &  \rightarrow	   \left.C\p{P,Z} \right) \\
    \land \left(N\p{P,H} \wedge E\p{H,Z}\right. &  \rightarrow \left.\neg C\p{P,Z}\right) 	 \\
    \land \left(N\p{P,H} \wedge C\p{H,Z}\right. &  \rightarrow \left.\neg E\p{P,Z}\right) \label{eq:triplewise}
    \end{split}
\end{align}

\subsection{Inconsistency Losses}
\label{sec:inconsistency-losses}

Using the consistency constraints stated in
\S\ref{sec:nli-learning-objective}, we can now derive the
inconsistency losses to minimize. For brevity, we will focus on the
annotation and symmetry consistencies.

First, let us examine annotation consistency. We can write the universal
quantifier in~\eqref{eq:labeled} as a conjunction to get:
\begin{align}
  \Land_{(P,H),Y^\star \in D} \top \rightarrow Y^\star{\p{P,H}} 
\end{align}
Using the product t-norm from Table~\ref{tab:tnorm}, we get the
learning objective of maximizing the probability of the true labels:
\begin{align}
  \prod_{\p{P,H},Y^\star\in D} y^\star_{\p{P,H}}  
\end{align}
Or equivalently, by transforming to the negative log space, we get the
annotation loss:
\begin{align}
  L_{ann} = \sum_{\p{P,H},Y^\star\in D} - \log y^\star_{\p{P,H}}.
\end{align}
We see that we get the familiar cross-entropy loss function using the definition
of inconsistency with the product t-norm\footnote{\citet{rocktaschel2015injecting} had a similar finding.}!

Next, let us look at symmetry consistency:
\begin{align}
\Land_{(P,H) \in D} C{\p{P,H}} \leftrightarrow C{\p{H,P}}.
\end{align}
Using the product t-norm, we get:
\begin{align}
  \prod_{\p{P,H}\in D} \min\p{ 1, \frac{c_{\p{H,P}}}{c_{\p{P,H}}}} & \min \p{1, \frac{c_{\p{P,H}}}{c_{\p{H,P}}}} \label{eq:pairwiseobj2}
\end{align}
 Transforming to the negative log space as before, we get the symmetry loss:
\begin{align}
  L_{sym} =  \sum_{\p{P,H}\in D} | \log c_{\p{P,H}} - & \log c_{\p{H,P}} | 
\end{align}

The loss for transitivity $L_{tran}$ can also be similarly derived.
We refer the reader to the appendix for details.

The important point is that we can systematically convert
logical statements to loss functions and cross-entropy is only one of such
losses.
To enforce some or all of these constraints, we
add their corresponding losses.
In our case study, with all constraints, the goal of learning is to minimize:
\begin{align}
  L &= L_{ann} + \lambda_{sym} L_{sym} + \lambda_{tran} L_{tran}
\end{align}
Here, the $\lambda$'s are hyperparameters to control the influence
of each loss term.

\subsection{Training Constrained Models}
\label{sec:nli-training}
The derived loss functions are directly compatible with off-the-shelf optimizers.
The symmetry/transitivity consistencies admit using unlabeled examples, 
while annotation consistency requires labeled examples.
Thus, in \S\ref{sec:experiments}, we will use both labeled and unlabeled data to power training.

Ideally, we want the unlabeled dataset to be absolutely informative, meaning a
model learns from \emph{every} example.  Unfortunately, obtaining such a dataset
remains an open question since new examples are required to be both linguistically meaningful
and difficult enough for the model.  \citet{minervini2018adversarially} used a language
model to generate unlabeled adversarial examples. Another way is via pivoting 
through a different language, which has a long history in machine
translation~\cite[\eg,][]{kay1997proper,mallinson2017paraphrasing}.

Since our focus is to study inconsistency, as an alternative, we propose a
simple method to create unlabeled examples: we randomly sample sentences from
the same topic.  In \S\ref{sec:experiments}, we will show that even random
sentences can be surprisingly informative because the derived losses operate in
real-valued space instead on discrete decisions.

\begin{table*}
  \centering
  \setlength{\tabcolsep}{4pt}
  \begin{tabular}{l|cccc|cccc}
    \toprule
    & \multicolumn{4}{c|}{5\%}& \multicolumn{4}{c}{100\%} \\
    Config & $\rho_{{S}}$ & $\tau_{{S}}$ & $\rho_{{T}}$ & $\tau_{{T}}$ & $\rho_{{S}}$ & $\tau_{{S}}$ & $\rho_{{T}}$ & $\tau_{{T}}$ \\
    \hline
    BERT w/ SNLI & 26.3 & 64.4 & 4.9 & 14.8 & 18.6 & 60.3 & 4.7 & 14.9 \\
    BERT w/ MultiNLI & 28.4 & 69.3 & 7.0 & 18.5 & 20.6 & 58.9 & 5.6 & 17.5 \\
    BERT w/ SNLI+MultiNLI & 25.3 & 62.4 & 4.8 & 14.8 & 18.1 & 59.6 & 4.5 & 14.8 \\
    BERT w/ SNLI+MultiNLI$^2$ & 22.1 & 67.1 & 4.1 & 13.7 & 19.3 & 59.7 & 4.5 & 15.2 \\
    \hline
    LSTM w/ SNLI+MultiNLI & 25.8 & 69.5 & 9.9 & 21.0 & 16.8 & 53.6 & 5.3 & 16.0 \\
    \bottomrule
  \end{tabular}
  \caption{Inconsistencies (\%) of models on our $100$k evaluation dataset.
  Each number represents the average of three random runs.
  Models are trained using $5$\% and $100$\% of the train sets.
  SNLI+MultiNLI$^2$: finetuned twice.
  $\rho_S$ and $\tau_S$: symmetry consistency violations.
  $\rho_T$ and $\tau_T$: transitivity consistency violations.}
  \label{tab:inconsistency}
\end{table*}



\section{Experiments}
\label{sec:experiments}

\begin{figure}
    \centering
    \includegraphics[width=\columnwidth]{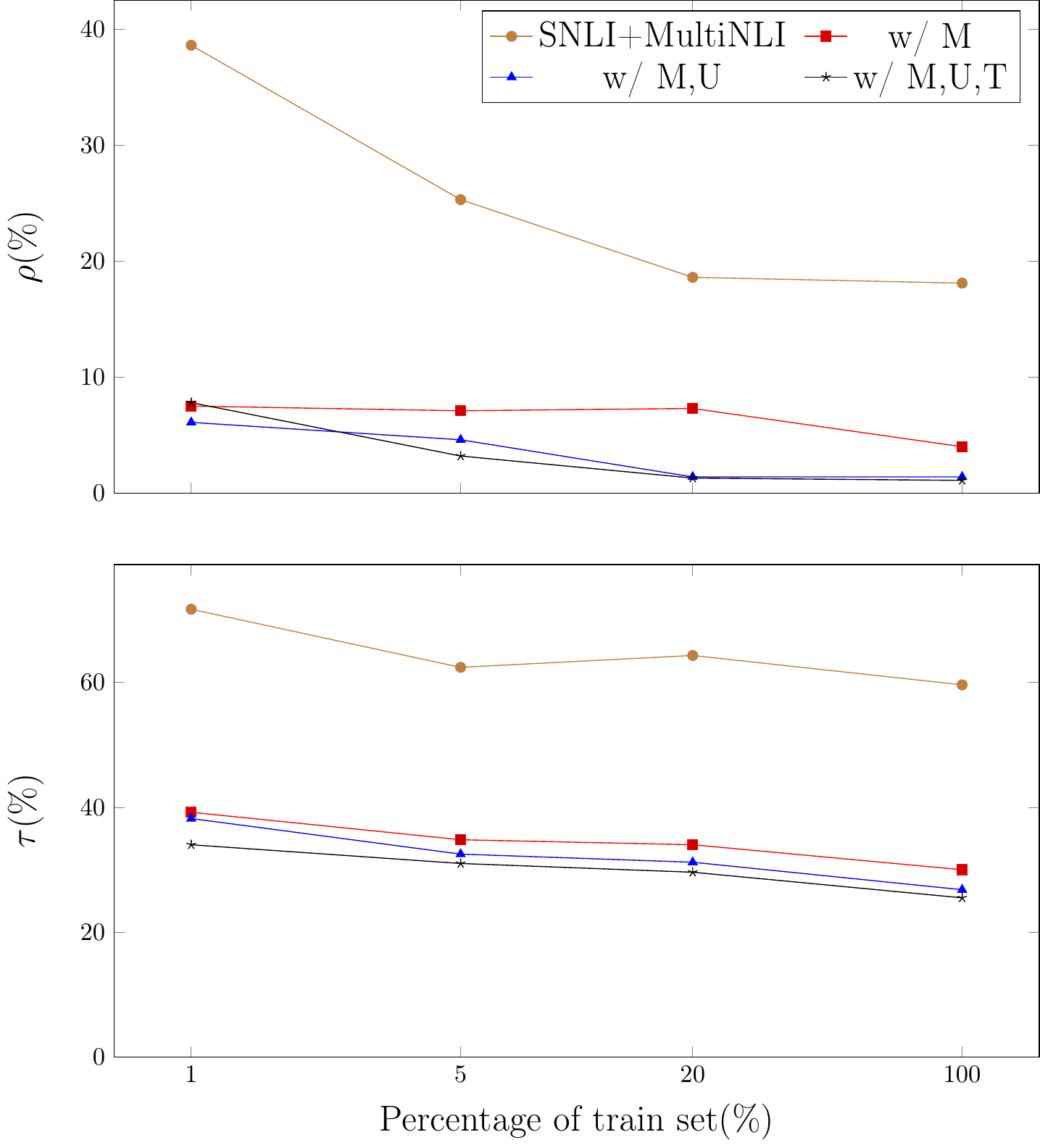}

    \caption{Symmetry inconsistencies on the $100$k evaluation example pairs.
    Each point represents the average of three random runs.
    M, U, and T are our unlabeled datasets with corresponding losses.}
    \label{fig:mirrorerr}
\end{figure}

\begin{figure}
  \centering
    \includegraphics[width=\columnwidth]{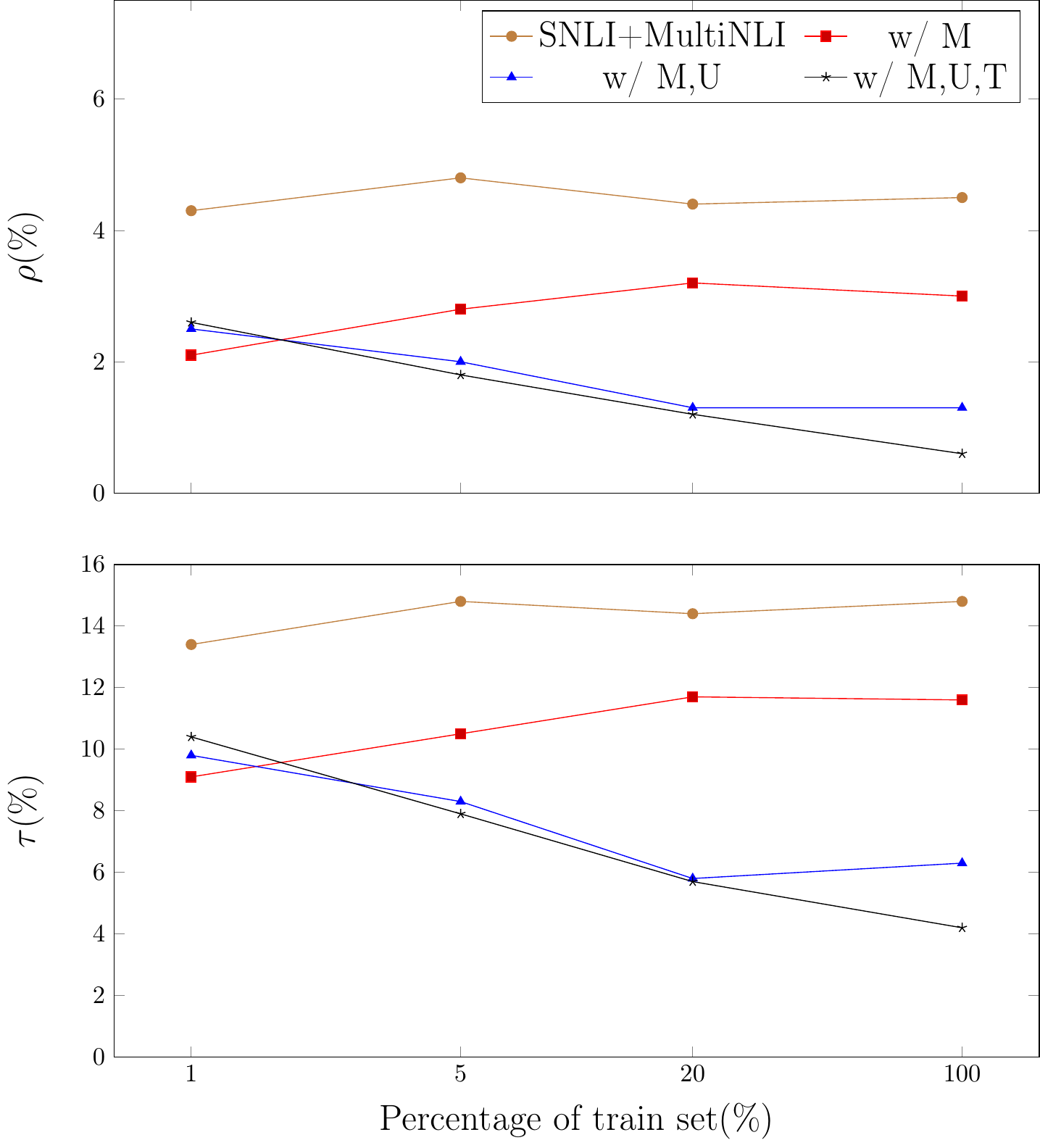}  
    \caption{Transitivity inconsistencies on the $100$k evaluation example pairs.
    Each point represents the average of three random runs.
    M, U, and T are our unlabeled datasets with corresponding losses.}
    \label{fig:transerr}
\end{figure}

In this section, we evaluate our framework using (near) state-of-the-art
approaches for NLI, primarily based on BERT, and also compare to an LSTM
model. We use the SNLI and
MultiNLI~\cite{wang2018glue} datasets to define annotation
consistency.  Our LSTM model is based on the decomposable attention
model with a BiLSTM encoder and GloVe embeddings~\cite{pennington2014glove}.  Our
BERT model is based on the pretrained BERT$_{base}$, finetuned on
SNLI/MultiNLI.  The constrained models are initialized with the
finetuned BERT$_{base}$ and finetuned again with inconsistency
losses.\footnote{This is critical when label supervision is
  limited.}  For fair comparison, we also show results of
BERT$_{base}$ models finetuned twice.

Our constrained models are trained on both labeled and unlabeled examples.
We expect that the different inconsistencies do not conflict with each other.
Hence, we select hyperparameters (\eg, the $\lambda$'s) using  development
accuracy only (\ie, annotation consistency).
We refer the reader to the appendix for details of our experimental
setup.

\subsection{Datasets}
To be comprehensive, we will use both of the SNLI and MultiNLI to train
our models, but we also show individual results.

We study the impact of the amount of label supervision by randomly sampling different percentages of labeled examples.
For each case, we also sample the same percentages from the corresponding development sets for model selection.
For the MultiNLI dataset, we use the {\tt matched} dev for
validation and {\tt mismatched} dev for evaluation.

\paragraph{Mirrored Instances (M)}
Given a labeled example, we construct its mirrored version by swapping the premise and the hypothesis.
This results in the same number of unlabeled sentence pairs as the annotated dataset.
When sampling by percentage, we will only use the sampled examples to construct mirrored examples.
We use this dataset for symmetry consistency.

\paragraph{Unlabeled Instance Triples (T)}
For the transitivity constraint, we sample $100$k sentence triples from MS~COCO~\cite{lin2014microsoft} captions.
From these, we construct three examples as in
\S\ref{sec:nli-learning-objective}: sentences $(P,H,Z)$ gives the pairs $(P,H)$, $(H,Z)$, and $(P,Z)$.
In all, we have $100$k example \emph{unlabeled} triples for the transitivity constraint.

\paragraph{Unlabeled Instance Pairs (U)}
For each sentence triple in the dataset T, we take the first example $(P,H)$ and construct mirrored examples, \ie $(H,P)$.
This yields $100$k \emph{unlabeled} instance pairs for training with
the symmetry loss.

\paragraph{Evaluation Dataset} We sample a different set of $100$k
example triples for measuring transitivity consistency.  For
symmetry consistency, we follow the above procedure for the
dataset U to construct evaluation instance pairs. Recall that the
definition of inconsistency allows measuring model quality with
unlabeled data.

\subsection{Inconsistency of Neural Models}
In Table~\ref{tab:inconsistency}, we report the impact of the amount
of annotated data on symmetry/transitivity consistencies by using different percentages of labeled examples.
We see that both LSTM and BERT models have symmetry consistency violations, while the transitivity consistency has lower violations.
Surprisingly, the LSTM model performed on par with BERT in terms of symmetry/transitivity
consistency; stronger representations does not necessarily mean more
consistent models.

The table shows that, given an example and its mirrored version, if the BERT
baseline predicts a \emph{Contradiction} on one, it has about $60$\% chance
($\tau_S$) to make an inconsistent judgement on the other.
Further, we see that the inconsistencies are not affected much by different datasets.
Models trained on the SNLI are as inconsistent as ones trained on MultiNLI.
Combining them only gives slight improvements.
Also, finetuning twice does not improve much over models finetuned once.

Finally, with more annotation, a model has fewer symmetry
consistency violations.
However, the same observation does not apply to the transitivity consistency.
In the following sections, we will show that we can almost
annihilate these inconsistencies using the losses from \S\ref{sec:inconsistency-losses}.

\begin{table*}
  \centering
  \setlength{\tabcolsep}{4pt}
  \begin{tabular}{l|cc|cc|cc|cc}
    \toprule
    & \multicolumn{2}{c|}{1\%}& \multicolumn{2}{c|}{5\%} & \multicolumn{2}{c|}{20\%}& \multicolumn{2}{c}{100\%} \\
    Config & SNLI & MultiNLI & SNLI & MultiNLI & SNLI & MultiNLI & SNLI & MultiNLI \\
    \hline
    SNLI+MultiNLI & 79.7 & 70.1 & 84.6 & 77.2 & 87.8 & 80.6 & 90.1 & 83.5 \\
    SNLI+MultiNLI$^2$ & 80.3 & 71.0 & 85.3 & 77.4 & 87.9 & 80.7 & 90.3 & 84.0 \\
    \hline
    w/ M & 80.1 & 71.0 & 85.3 & 77.8 & 88.1 & 80.6 & 90.3 & 84.1 \\
    w/ M,U & 80.2 & 71.0 & 85.4 & 77.2 & 88.1 & 80.9 & 90.5 & 84.3 \\
    w/ M,U,T & 80.6 & 71.1 & 85.4 & 77.2 & 88.1 & 80.9 & 90.2 & 84.2 \\
    \bottomrule
  \end{tabular}
  \caption{Impact of symmetry/transitivity consistencies on test set accuracies.
  Each number represents the average of three random runs of BERT$_{base}$.
  Columns are accuracies on our SNLI/MultiNLI test sets.
  SNLI+MultiNLI$^2$: finetuned twice.
  M, U, and T are unlabeled datasets with respective inconsistency losses.}
  \label{tab:impact}
\end{table*}

\subsection{Reducing Inconsistencies}
We will study the effect of symmetry and transitivity consistency losses
in turn using the BERT models.
To the baseline models, we incrementally include the M, U, and T datasets.
We expect that the constrained models should have accuracies
at least on par with the baseline (though one of the key points of this paper is that accuracy by itself is not a comprehensive metric).

In Fig.~\ref{fig:mirrorerr}, we present both of the global and conditional violation rates of baselines and the constrained models.
We see that mirrored examples (\ie, the w/ M curve) greatly reduced the symmetry inconsistency.
Further, with $100$k unlabeled example pairs (the w/ M,U curve), we can further reduce the error rate.
The same observation also applies when combining symmetry with transitivity constraint.

Fig.~\ref{fig:transerr} shows the results for transitivity inconsistency.
The transitivity loss is, again, greatly reduced both for the global and conditional violations.
We refer the reader to the appendix for exact numbers.

We see that with our augmented losses, even a model using $1$\% label supervision can be much more consistent than the baselines trained on $100$\% training set!
This suggests that label supervision does not explicitly encode the
notion of consistency, and consequently models do not get this
information from the training data.

With the simultaneous decline in global and conditional violation
rate, the constrained models learn to agree with the
consistency requirements specified declaratively.
We will see in the next section, doing so does not sacrifice model accuracies.

\subsection{Interaction of Losses}
In Table~\ref{tab:impact}, we show the impact of symmetry and transitivity consistency on test accuracy.
And the interaction between symmetry and transitivity consistency is covered in Fig~\ref{fig:mirrorerr} and~\ref{fig:transerr}.

Our goal is to minimize all inconsistencies without sacrificing one for another.
In Table~\ref{tab:impact}, we see that lower symmetry/transitivity
inconsistency generally does not reduce test accuracy, but we do not observe substantial improvement either.
In conjunction with the observations from above, this
suggests that test sets do not explicitly measure symmetry/transitivity consistency.

From Fig~\ref{fig:mirrorerr} and~\ref{fig:transerr}, we see that models constrained by both symmetry and transitivity losses are generally more consistent than models using symmetry loss alone.
Further, we see that in Fig.~\ref{fig:transerr}, using mirrored dataset alone can even mitigate the transitivity errors.
With dataset P, the transitivity inconsistency is strongly reduced by the symmetry inconsistency loss.
These observations suggest that the compositionality of constraints does not pose internal conflict to the model.
They are in fact beneficial to each other.

Interestingly, in Fig~\ref{fig:transerr}, the models trained with mirrored dataset (w/ M) become more inconsistent in transitivity measurement when using more training data. We believe there are two factors causing this. Firstly, there is a vocabulary gap between SNLI/MultiNLI data and our unlabeled datasets (U and T). Secondly, the w/ M models are trained with symmetry consistency but evaluated with transitivity consistency. The slightly rising inconsistency implies that, without vocabulary coverage, training with one consistency might not always benefit another consistency, even using more training data.

When label supervision is limited (\ie $1$\%), the models can easily overfit via the transitivity loss.
As a result, models trained on the combined losses (\ie w/ M,U,T) have slightly larger transitivity inconsistency than models trained with mirrored data (\ie w/ M) alone.
In fact, if we use no label supervision at all, the symmetry and transitivity losses can push every prediction towards label \emph{Neutral}.
But such predictions sacrifice annotation consistency.
Therefore, we believe that some amount of label supervision is necessary.


\section{Analysis}
\label{sec:analysis}
In this section, we present an analysis of how the different losses affect model prediction and how informative they are during training.

\subsection{Coverage of Unlabeled Dataset}
Table~\ref{tab:coverage} shows the {\em coverage} of the three unlabeled datasets during the first training epoch.
Specifically, we count the percentage of unlabeled examples where the symmetry/transitivity loss is positive.
The coverage decreases in subsequent epochs as the
model learns to minimize constraint violations.
We see that both datasets M and U have high coverage.
This is because that, as mentioned in \S\ref{sec:framework}, our loss function works in real-valued space instead of discrete decisions.
The coverage of the dataset T is much lower because the compositional antecedent in transitivity
statements holds less often, which naturally leads to smaller
coverage, unlike the unary antecedent for symmetry.

\begin{table}[t]
  \centering
  \setlength{\tabcolsep}{4pt}
  \begin{tabular}{l|ccc}
    \toprule
    Data & M & U & T \\
    \hline
    5\% w/ M,U,T & 99.8 & 99.4 & 12.0 \\
    100\% w/ M,U,T & 98.7 & 97.6 & 6.8 \\
    \bottomrule
  \end{tabular}
  \caption{Coverage (\%) of unlabeled training sentences during the first epoch of training.
  Percentages are calculated from models with random seed 1.}
  \label{tab:coverage}
\end{table}

\subsection{Distribution of Predictions}
In Table~\ref{tab:pairdist}, we present the distribution of model predictions on the $100$k evaluation example pairs for symmetry consistency.
Clearly, the number of constraint-violating (off-diagonal) predictions significantly dropped.
Also note that the number of \emph{Neutral} nearly doubled in our constrained model.
This meets our expectation because the example pairs are constructed from randomly sampled sentences under the same topic.

We also present the distribution of predictions on example triples for the transitivity consistency in Table~\ref{tab:tripledist}.
As expected, with our transitivity consistency, the distribution
of the label \emph{Neutral} gets significantly higher as well.
Further, in Table~\ref{tab:tripleinderr}, we show the error rates of each individual transitivity consistencies.
Clearly our framework mitigated the violation rates on all four statements.

While the logic-derived regularization pushes model prediction on unlabeled
datasets towards \emph{Neutral}, the accuracies on labeled test sets are not
compromised. We believe this relates to the design of current NLI datasets where
the three labels are balanced. But in the real world, neutrality represents
potentially infinite negative space while entailments and contradictions are rarer.
The total number of neutral examples across both the SNLI and MultiNLI test sets is about 7k.
\emph{Can we use these $7$k examples to evaluate the nearly infinite negative space?} We believe not.

\begin{table}[t]
  \centering
  \setlength{\tabcolsep}{2.5pt}
  \begin{tabular}{rr|rrr|rrr}
    \toprule
    & & \multicolumn{3}{c|}{BERT} & \multicolumn{3}{c}{w/ M,U,T}  \\
    & & \multicolumn{3}{c|}{$(H, P)$} & \multicolumn{3}{c}{$(H, P)$} \\
    &  & E & C & N & E & C & N \\
    \hline
    \multirow{3}{*}{\rotatebox[origin=c]{90}{$(P, H)$}} & E & 4649 & \textbf{1491} & 14708 & 2036 & \textbf{29} & 9580 \\
    & C & \textbf{1508} & 10712  & \textbf{6459} & \textbf{33} & 4025  & \textbf{627} \\
    & N & 14609 & \textbf{6633} & 39231 & 9632 & \textbf{613} & 73425 \\
    \bottomrule
  \end{tabular}
  \caption{Distribution of predictions on the $100$k evaluation
    data using BERT 
  trained on $100\%$ SNLI+MultiNLI data with random seed 1.
  Bold entries are symmetrically inconsistent.}
  \label{tab:pairdist}
\end{table}

\begin{table}[t]
  \centering
  \setlength{\tabcolsep}{4pt}
  \begin{tabular}{lc|ccc}
    \toprule
    Model & Example & E & C & N \\
    \hline
    \multirow{3}{*}{\rotatebox[origin=c]{0}{BERT}} & $(P, H)$ & 20848 & 18679 & 60473 \\
    & $(H, Z)$ & 20919 & 18768 & 60313 \\
    & $(P, Z)$ & 20779 & 18721 & 60500 \\
    \hline
    \multirow{3}{*}{\rotatebox[origin=c]{0}{w/ M,U,T}} & $(P, H)$ & 11645 & 4685 & 83670 \\
    & $(H, Z)$ & 11671 & 4703 & 83626 \\
    & $(P, Z)$ & 11585 & 4597 & 83818 \\
    \bottomrule
  \end{tabular}
  \caption{Distribution of predictions on the $100$k evaluation example triples.
  BERT: trained on the full SNLI+MultiNLI data.
  Predictions are from random run with seed 1.}
  \label{tab:tripledist}
\end{table}

\begin{table}[t]
  \centering
  \setlength{\tabcolsep}{4pt}
  \begin{tabular}{l|cc|cc}
    \toprule
    & \multicolumn{2}{c|}{BERT} & \multicolumn{2}{c}{w/ M,U,T} \\
    Transitivity & $\rho_T$ & $\tau_T$ & $\rho_T$ & $\tau_T$ \\
    \hline
    $E\wedge E \rightarrow E$ & 0.7 & 16.0 & 0.2 & 15.1 \\
    $E\wedge C \rightarrow C$ & 1.8 & 49.6 & 0.2 & 46.5 \\
    $N\wedge E \rightarrow \neg C$ & 1.2 & 9.0 & 0.2 & 1.8 \\
    $N\wedge C \rightarrow \neg E$ & 1.0 & 9.3 & 0.1 & 4.8 \\
    \bottomrule
  \end{tabular}
  \caption{Individual transitivity inconsistency (\%) on the $100$k evaluation example triples.
  BERT: trained on the full SNLI+MultiNLI data.
  Predictions are from random run with seed 1.}
  \label{tab:tripleinderr}
\end{table}


\section{Related Works and Discussion}
\label{sec:related}

\paragraph{Logic, Knowledge and Statistical Models}
Using soft relaxations of Boolean formulas as loss functions has rich history in
AI.  The \L{}ukasiewicz t-norm drives knowledge-driven learning and inference in
probabilistic soft logic~\cite{kimmig2012short}. \citet{li2019augmenting} show
how to augment existing neural network architectures with domain knowledge using
the \L{}ukasiewicz t-norm. \citet{pmlr-v80-xu18h} proposed a general framework
for designing a semantically informed loss, without t-norms, for constraining a
complex output space.  In the same vein, \citet{fischer2019dl2} also proposed a
framework for designing losses with logic, but using a bespoke mapping of the
Boolean operators.

Our work is also conceptually related to posterior
regularization~\cite{ganchev2010posterior} and constrained conditional
models~\cite{chang2012structured}, which integrate knowledge with statistical
models.  Using posterior regularization with imitation learning,
\citet{hu2016harnessing} transferred knowledge from rules into neural
parameters.  \citet{rocktaschel2015injecting} embedded logic into distributed
representations for entity relation extraction.  \citet{alberti2019synthetic} imposed
answer consistency over generated questions for machine comprehension.
Ad-hoc regularizers have been
proposed for process comprehension~\cite{du2019consistent}, semantic role
labeling~\cite{mehta2018towards}, and summarization~\cite{hsu2018unified}.

\paragraph{Natural Language Inference} 
In the literature, it has been shown that even highly accurate models show a
decline in performance with perturbed examples.  This lack of robustness of NLI
models has been shown by comparing model performance on pre-defined
propositional rules for swapped datasets~\cite{wang2018if} or outlining
large-scale stress tests to measure stability of models to semantic, lexical and
random perturbations~\cite{naik2018stress}. Moreover, adversarial training
examples produced by paraphrasing training data~\cite{iyyer2018adversarial} or
inserting additional seemingly important, yet unrelated, information to training
instances~\cite{jia2017adversarial} have been used to show model inconsistency.
Finally, adversarially labeled examples have been shown to improve prediction
accuracy~\cite{kang2018adventure} . Also related in this vein is the idea of
dataset inoculation~\cite{liu2019inoculation}, where models are finetuned by
exposing them to a challenging dataset.

The closest related work to this paper is probably that
of~\citet{minervini2018adversarially}, which uses the G\"odel t-norm to discover
adversarial examples that violate constraints.  There are three major
differences compared to this paper:
\begin{inparaenum}[1)]
\item our definition of inconsistency is a strict generalization of errors of
  model predictions, giving us a unified framework for that includes
  cross-entropy as a special case,
\item our framework does not rely on the construction of adversarial datasets,
  and
\item we studied the interaction of annotated examples vs. unlabeled examples via constraint, showing that our constraints can yield strongly consistent model with even a small amount of label supervision.
\end{inparaenum}



\section{Conclusion}

In this paper, we proposed a general framework to measure and mitigate model inconsistencies.
Our framework systematically derives loss functions from domain knowledge stated in logic rules to constrain model training.
As a case study, we instantiated the framework on a state-of-the-art model for the NLI task, showing that models can be highly accurate and consistent at the same time.
Our framework is easily extensible to other domains with rich output structure, \eg, entity relation extraction, and multilabel classification.


\section*{Acknowledgements}
We thank members of the NLP group at the University of Utah for
their valuable insights and suggestions, especially Mattia Medina Grespan for pointing out $R$-fuzzy logic and $S$-fuzzy logic; and reviewers for pointers
to related works, corrections, and helpful comments. We also
acknowledge the support of NSF SaTC-1801446, and gifts from Google
and NVIDIA.


\bibliographystyle{style/acl_natbib}
\bibliography{cited}

\appendix
\section{Appendices}
\label{sec:appendix}

\subsection{Violations as Generalizing Errors}

Both global and conditional violations defined in the body of the
paper generalize classifier error. In this section, we will show
that for a dataset with only labeled examples, and no additional
constraints, both are identical to error.

Recall that an example $x$ annotated with label $Y^\star$ can be
written as $\top \rightarrow Y^\star(x)$. If we have a dataset $D$
of such examples and no constraints, in  our unified representation
of examples, we can write this as the following conjunction:
\begin{align*}
  \forall x\in D,\quad \top \rightarrow Y^\star(x).
\end{align*}
We can now evaluate the two definitions of violation for this
dataset.

First, note that the denominator in the definition of the
conditional violation $\tau$ counts the number of examples because
the antecedent for all examples is always true. This makes $\rho$
and $\tau$ equal. Moreover, the numerator is the number of examples
where the label for an example is not $Y^\star$. In other words, the
value of $\rho$ and $\tau$ represents the fraction of examples in
$D$ that are mislabeled.

The strength of the unified representation and the definition of
violation comes from the fact that they apply to arbitrary
constraints.

\subsection{Loss for Transitivity Consistency}

This section shows the loss associated with the transitivity
consistency in the NLI case study. For an individual example
$(P, H, Z)$, applying the product t-norm to the definition of the
transitivity consistency constraint, we get the loss
\begin{align}\small
  \begin{split}
    &\text{\small ReLU}\p{\log e{\p{P,H}} \text{\small$+$} \log e{\p{H,Z}} \text{\small$-$} \log e{\p{P,Z}}} \\
    +&\text{\small ReLU}\p{\log e{\p{P,H}} \text{\small$+$} \log c{\p{H,Z}} \text{\small$-$} \log c{\p{P,Z}}} \\
    +&\text{\small ReLU}\p{\log n{\p{P,H}} \text{\small$+$} \log e{\p{H,Z}} \text{\small$-$} \log \p{1 \text{\small$-$} c{\p{P,Z}}}} \\
    +&\text{\small ReLU}\p{\log n{\p{P,H}} \text{\small$+$} \log c{\p{H,Z}} \text{\small$-$} \log \p{1 \text{\small$-$} e{\p{P,Z}}}} \label{eq:triplewiseobj}
  \end{split}
\end{align}
That is, the total transitivity loss $L_{tran}$ is the sum of this expression
over the entire dataset.

\subsection{Details of Experiments}

\begin{table*}
  \centering
  \setlength{\tabcolsep}{4pt}
  \begin{tabular}{l|cc|cc|cc|cc|cc|cc}
    \toprule
    & \multicolumn{6}{c|}{1\%}& \multicolumn{6}{c}{5\%} \\
    Train & SNLI & MultiNLI & $\rho_S$ & $\tau_S$ & $\rho_T$ & $\tau_T$ & SNLI & MultiNLI & $\rho_S$ & $\tau_S$ & $\rho_T$ & $\tau_T$ \\
    \hline
    SNLI & 79.3 & na & 36.7 & 70.6 & 6.1 & 17.1 & 84.5 & na & 26.3 & 64.4 & 4.9 & 14.8 \\
    MultiNLI & na & 69.0 & 29.1 & 83.1 & 8.2 & 18.4 & na & 76.1 & 28.4 & 69.3 & 7.0 & 18.5 \\
    SNLI+MultiNLI & 79.7 & 70.1 & 38.6 & 71.7 & 4.3 & 13.4 & 84.6 & 77.2 & 25.3 & 62.4 & 4.8 & 14.8 \\
    SNLI+MultiNLI$^2$ & 80.3 & 71.0 & 32.4 & 75.0 & 3.9 & 12.8 & 85.3 & 77.4 & 22.1 & 67.1 & 4.1 & 13.7 \\
    \hline
    w/ M   & 80.1 & 71.0 & 7.5 & 39.2 & 2.1 & 9.1 & 85.3 & 76.8 & 7.1 & 34.8 & 2.8 & 10.5 \\
    w/ M,U & 80.2 & 71.0 & 6.1 & 38.2 & 2.5 & 9.8 &  85.4 & 77.2 & 4.6 & 32.5 & 2.0 & 8.3 \\
    w/ M,U,T & 80.6 & 71.1 & 7.8 & 34.0 & 2.6 & 10.4 & 85.4 & 77.2 & 3.2 & 31.0 & 1.8 & 7.9 \\
    \hline\hline
    & \multicolumn{6}{c|}{20\%}& \multicolumn{6}{c}{100\%} \\
    Train & SNLI & MultiNLI & $\rho_S$ & $\tau_S$ & $\rho_T$ & $\tau_T$ & SNLI & MultiNLI & $\rho_S$ & $\tau_S$ & $\rho_T$ & $\tau_T$ \\
    \hline
    SNLI & 87.5 & na & 21.2 & 63.0 & 4.1 & 13.6 & 90.1 & na & 18.6 & 60.3 & 4.7 & 14.9 \\
    MultiNLI & na & 80.4 & 25.8 & 58.1 & 5.1 & 16.5 & na & 83.7 & 20.6 & 58.9 & 5.6 & 17.5 \\
    SNLI+MultiNLI & 87.8 & 80.6 & 18.6 & 64.3 & 4.4 & 14.4 & 90.1 & 83.5 & 18.1 & 59.6 & 4.5 & 14.8 \\
    SNLI+MultiNLI$^2$ & 87.9 & 80.7 & 19.0 & 64.0 & 4.3 & 14.5 & 90.3 & 84.0 & 19.3 & 59.7 & 4.5 & 15.2\\
    \hline
    w/ M   & 88.1 & 80.6 & 7.3 & 34.0 & 3.2 & 11.7 & 90.3 & 84.1 & 6.2 & 28.1 & 3.0 & 11.6 \\
    w/ M,U & 88.1 & 80.9 & 1.4 & 31.2 & 1.3 & 5.8 & 90.5 & 84.3 & 1.4 & 26.8 & 1.3 & 6.3 \\
    w/ M,U,T & 88.1 & 80.9 & 1.3 & 29.6 & 1.2 & 5.7 & 90.2 & 84.2 & 1.1 & 25.5 & 0.6 & 4.2 \\
    \bottomrule
  \end{tabular}
  \caption{Symmetry/Transitivity inconsistencies (\%) for models using $1$\%, $5$\%, $20$\%, and $100$\% training data.
  Each number represents the average of three random runs.
  SNLI+MultiNLI$^2$: BERT$_{base}$ finetuned twice for fair comparison.
  SNLI/MultiNLI column: accuracies on corresponding text sets.
  M: mirrored labeled examples.
  U: unlabeled instance pairs.
  T: unlabeled instance triples.}
  \label{tab:1to100}
\end{table*}

\subsubsection{Setup}

For BERT$_{base}$ baselines, we finetune them for $3$ epochs with learning rate $3\times 10^{-5}$, warmed up for all gradient updates.
For constrained models, we further finetune them for another $3$ epochs with lowered learning rate $1\times 10^{-5}$.
When dataset $U$ is present, we further lower the learning rate to $5\times 10^{-6}$.
Optimizer is Adam across all runs.
During training, we adopt Dropout rate~\cite{srivastava2014dropout} $0.1$ inside of BERT transformer encoder while $0$ at the final linear layer of classification.

For different types of data and different consistency constraints, we used different weighting factors $\lambda$`s.
In general, we found that the smaller amount of labeled examples, the smaller $\lambda$ for the symmetry and transitivity consistency.
In Table~\ref{tab:choiceoflambd}, we see that the $\lambda$`s for U and T grows exponentially with the size of annotated examples.
In contrast, the $\lambda$ for M dataset can be much higher. We found a good value for M is $1$.
This is because the size of dataset $U$ and $T$ are fixed to be $100$k, while the size of dataset $M$ is the same as the amount of labeled examples.

Having larger $\lambda$ leads to significantly worse accuracy on the development set, especially that of SNLI.
Therefore we did not select such models for evaluation.
We hypothesize that it is because the SNLI and MultiNLI are crowdsourced from different domains while the MS COCO shares the same domain as the SNLI.
Larger scaling factor could push unlabeled examples towards \emph{Neutral}, thus sacrificing the annotation consistency on SNLI examples.

\begin{table}
  \centering
  \setlength{\tabcolsep}{4pt}
  \begin{tabular}{l|cccc}
    \toprule
    Data & $1$\% & $5$\% & $20$\% & $100$\% \\
    \hline
    M & $1$ & $1$ & $1$ & $1$ \\
    U & $10^{-5}$ & $10^{-4}$ & $10^{-3}$ & $10^{-1}$ \\
    T & $10^{-6}$ & $10^{-5}$ & $10^{-4}$ & $10^{-3}$ \\
    \bottomrule
  \end{tabular}
  \caption{Choice of $\lambda$`s for different consistency and corresponding unlabeled datasets.
  For different sizes of annotation and different types of data, we adopt different $\lambda$`s.}
  \label{tab:choiceoflambd}
\end{table}

\subsubsection{Results}
We present the full experiment results on the natural language inference task in Table~\ref{tab:1to100}.
Note that the accuracies of baselines finetuned twice are slightly better than models only finetuned once,
while their symmetry/transitivity consistencies are roughly on par.
We found such observation is consistent with different finetuning hyperparameters (\eg warming, epochs, learning rate).


\end{document}